\documentclass[10pt,twocolumn,letterpaper]{article}

\usepackage[pagenumbers]{cvpr} % To force page numbers, e.g. for an arXiv version

\definecolor{cvprblue}{rgb}{0.21,0.49,0.74}
\usepackage[pagebackref,breaklinks,colorlinks,allcolors=cvprblue]{hyperref}
\usepackage{pifont} % for checkmark and cross symbols
\usepackage{pifont}       % for checkmarks and crosses
\usepackage{booktabs}     % professional rules
\usepackage{multirow}     % multirow support
\usepackage{colortbl}     % for row colors
\usepackage{xcolor}       % for custom colors
\usepackage[skip=2pt]{caption} % default is 10pt
\usepackage{multirow}
\usepackage{amsmath}
\usepackage{amssymb}
\usepackage{booktabs}
\usepackage{graphicx}
\usepackage{adjustbox}
\usepackage[margin=1in]{geometry}
\usepackage{booktabs}
\usepackage{caption}
\usepackage{algorithm}
\usepackage{algpseudocode}
\usepackage{booktabs}
\usepackage{afterpage}
\def\paperID{-----} % *** Enter the Paper ID here
\def\confName{CVPR}
\def\confYear{2026}
\title{ Cranio-ID: Graph-Based Craniofacial Identification via Automatic Landmark Annotation in 2D Multi-View X-rays}
\author{
Ravi Shankar Prasad$^{1}$ \quad
Nandani Sharma$^{1}$ \quad
Dinesh Singh$^{1}$\\[4pt]
$^{1}$Visual Intelligence and Machine Learning (VIML) Group,\\
School of Computing and Electrical Engineering,\\
Indian Institute of Technology Mandi, India\\[4pt]
{\tt\small \{d23033, d22180\}@students.iitmandi.ac.in,
dineshsingh@iitmandi.ac.in}
}

\begin{document}
\maketitle

\begin{abstract}
In forensic craniofacial identification and in many biomedical applications, craniometric landmarks are important. Traditional methods for locating landmarks are time-consuming and require specialized knowledge and expertise. Current methods utilize superimposition and deep learning-based methods that employ automatic annotation of landmarks. However, these methods are not reliable due to insufficient large-scale validation studies. In this paper, we proposed a novel framework Cranio-ID: First, an automatic annotation of landmarks on  2D skulls (which are X-ray scans of faces) with their respective optical images using our trained YOLO-pose models. Second, cross-modal matching by formulating these landmarks into graph representations and then finding semantic correspondence between graphs of these two modalities using cross-attention and optimal transport framework. Our proposed framework is validated on the S2F and CUHK datasets (CUHK dataset resembles with S2F dataset). Extensive experiments have been conducted to evaluate the performance of our proposed framework, which demonstrates significant improvements in both reliability and accuracy, as well as its effectiveness in cross-domain skull-to-face and sketch-to-face matching in forensic science.
\end{abstract}
    
\section{Introduction}
\label{sec:intro}

Forensic craniofacial identification (FCI) aims to identify individuals on the basis of an unknown skull. Several studies~\cite{damas2020handbook},~\cite{damas2011forensic}~\cite{prasad2025cross},~\cite{bermejo2021automatic},~\cite{yun2022semi} have been conducted on FCI, which state the challenge and need for FCI in forensic sciences as well as in real-world applications. Traditional methods~\cite{hwang2012facial},~\cite{wilkinson2010facial},~\cite{fagertun20143d} for FCI, which include identification of the skull using manual annotation of landmarks on the skull or face, are challenging and can take a considerable amount of time, as they demand a high level of specialised skill. Recent methods on FCI use superimposition~\cite{damas2011forensic},~\cite{guerra2025international},~\cite{jayaprakash2015conceptual},~\cite{lenza2015radiographic} and deep learning-based approaches, but their reliability is questionable due to lack of extensive validation studies. Additionally, due to the absence of a publicly available pair-wise skull-to-face dataset, the study of FCI becomes more challenging, hindering effective comparisons between methods.
\begin{figure}[!t]
    \centering
    \includegraphics[width=1\linewidth, keepaspectratio,trim={9.39cm 4.4cm 7.8cm 3.2cm},clip]
                     {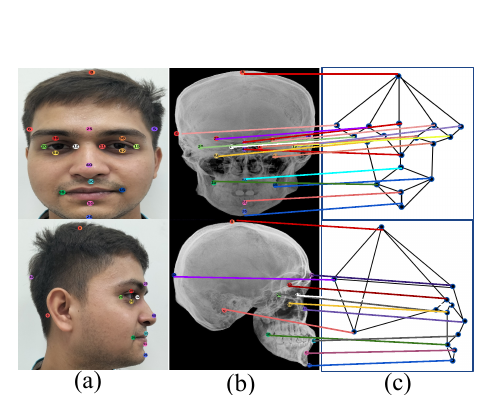}
    \caption{Sample image showing landmark localization on the face (a) and skull (b); where (b) and (c) show semantic correspondence between landmarks on skull and graph skeletons from two views: Front and Side. Total 18 landmarks are localized for the front and 13 for the side face and skull image respectively.}
    \label{fig:dataset_1}
\end{figure}
% \begin{figure}
%     \centering
%     \includegraphics[width =0.6\textwidth, keepaspectratio,trim={8cm 3cm 3cm 0cm},clip]{Images/graph_annotation.pdf}
%     \caption{ Sample image showing landmark graph skeleton on skull and face from two views, respectively.}
%     \label{fig:dataset_1}
% \end{figure}

% \begin{figure}
%     \centering
%     \includegraphics[width =0.5\textwidth, keepaspectratio,trim={3cm 1.5cm 1cm 0cm},clip]{Images/skull_face_landmarks.pdf}
%     % \caption{ Sample image showing landmark localization on skull and face from two views: Front and Side by YOLOv8 model. Total 18 landmarks are localized for the front and 13 for the side skull and face image respectively.}
%     \label{fig:dataset_1}
% \end{figure}

% Craniofacial landmarks localisation aims to put landmarks on the specific reference points that are located on the bone surfaces (known as craniometric analysis) or the soft-tissue surfaces (referred to as capulometric analysis). Hence, the precise identification of anatomical structure is important for craniofacial analysis. 

These landmark-based approaches are essential in a variety of professional fields, including medicine, dentistry, and forensic anthropology. It serves critical functions in several applications, such as planning craniofacial surgeries~\cite{kolar1997craniofacial}, conducting orthodontic assessments~\cite{downs1948variations}, ~\cite{kim2006three}, performing morphometric studies~\cite{ross1999allocation} that analyze shapes and forms, and aiding in the identification of individuals~\cite{damas2020handbook}, ~\cite{huang2011weighted} for forensic investigations. Hence, accurate, fast and reliable methods for FCI and automatic annotation of landmarks on skull and face are important.

 According to the study proposed by Bookstein~\cite{bookstein1997morphometric}, all landmarks are not uniformly recognizable, and their identification can vary significantly, because these landmarks have different geometric locations on the cranial structures.
 Hence, in this paper, we considering only those significant landmarks by which a skull and face of a person can be represented, our model localizes and detects total 19 landmarks, and the information of these landmarks are mentioned in supplementary. Figure~\ref{fig:dataset_1} illustrates these landmarks and semantic correspondence between frontal and lateral views of 2D skull and face images with their graph skeleton. There are several studies conducted on automatic facial landmark detection and localization, but very few are done with skull images. Although several works have been done on 3D skull images, to the best of our knowledge, few work has been conducted on 2D skull images (X-ray image of face). As we also want to match the skull with its corresponding face image, hence for this we use the architecture in~\cite{Sharma2025ParaX},~\cite{sharma2025exp} to solve our skull-to-face matching problem.
 Hence, to overcome all these challenges and problems, our research first focused on reducing the challenges faced by manual process, by providing an model which can automatically annotate or localize these landmarks on the cranial structures. Second, our proposed a framework (Cranio-ID) matches the skull and their corresponding face image by establishing semantics correspondence between two different modalities by leveraging combined cross-attention~\cite{wei2020multi} and optimal transport ~\cite{liu2020semantic}.
In summary, following are the contributions of this research;
\begin{itemize}

    \item We proposed a novel framework Cranio-ID, in which we trained a YOLO-pose model to automatically detect anatomical landmarks on 2D skull-to-face images.
    \item We demonstrate how cross-attention and optimal transport methods effectively establish semantic correspondence between the two modalities through extensive comparative experiments. 
    \item We conduct a comprehensive evaluation of the proposed framework for automatic landmark localization and skull-to-face and sketch-to-face matching, using metrics such as recall@k and mAP@k on two public datasets S2F~\cite{prasad2025cross} (skull-to-face) and CUHK~\cite{zhang2011coupled} (sketch-to-face).

\end{itemize}

% The rest of the paper is organized as follows: In section 2 we discussed the related work related to the detection of craniofacial landmarks. In section 3, we describe the preprocessing part which is performed on our benchmark dataset containing the skull and its respective face, followed by the YOLO model for detection of landmarks and its architecture. Section 4 includes details of our benchmark dataset. Section 5 presents our experimental validation and results. Finally, in section 6, we conclude the paper with the defined future direction. 

\section{Related work}
% \textbf{Craniofacial identification.}
In the identification of skulls, limited work has been done. Traditionally, research on skull identification mainly focuses on two types of categories. One is craniofacial superimposition~\cite{damas2011forensic},~\cite{damas2020handbook}, and the other is craniofacial reconstruction~\cite{claes2010bayesian},~\cite{claes2010computerized},~\cite{duan2014craniofacial},~\cite{hu2013hierarchical},~\cite{missal2023forensic},~\cite{paysan2009face},~\cite{vandermeulen2006computerized}. The work done in~\cite{damas2011forensic},~\cite{damas2020handbook} uses digital tools such as Skeleton-ID$^{TM}$~\footnote{http://skeleton-id.com/} to superimpose the skull onto the face. Hence, due to approximate overlapping or superimposition, these methods are not accurate and faithful, and the studies done in forensic reconstruction are mainly on 3D CB-CT data, which are costly and very difficult to get. Few works have been done for craniofacial identification based on landmarks~\cite{desvignes20063d},~\cite{huang2011weighted},~\cite{tu2007automatic}. However, all these methods struggle with limited paired skull-face data, as these studies were conducted on 3D CB-CT data, which is again difficult to obtain and process.
In conventional GCN-based methods~\cite{zhao2021geometry, zhang2014facial, zhao2022spatial, jin2024transformer, xu2024joint, liu2024descriptive, dong2024attentional, al2024ter, mao2025facial, qu2025design, huang2025modeling, tanchotsrinon2011facial, kassab2024gcf, xu2020facial, chen2024dual, ngoc2020facial}, the landmarks, action units (AUs), and their connections are usually defined beforehand, resulting in a graph structure that stays unchanged throughout training. Unlike these fixed designs, our model adopts a more adaptive and dynamic strategy. However, graph structures are highly effective at capturing structural information within an image, as they can represent both nearby pixels and distant regions, thereby highlighting how different parts of an image are connected and related to each other.

Sharma \textit{et al.}~\cite{sharma2025exp} proposed, Exp-Graph introduces a graph-based framework for facial expression recognition, where landmarks serve as vertices and edges capture both spatial proximity and appearance similarity. By combining vision transformers with graph convolutional networks, it effectively models local and global dependencies to improve recognition accuracy. Also proposed Para-X~\cite{Sharma2025ParaX}  employs a graph-based representation of facial attributes, where key-points form vertices and edges capture spatial and appearance relations. By integrating vision transformers with graph convolutional networks, it effectively models local–global dependencies for improved facial paralysis recognition. Several works have explored graph-based approaches for image classification, yet our model uses~\cite{Sharma2025ParaX,sharma2025exp} a more flexible and dynamic approach.

% \textbf{Cross-domain matching.}
Several works~\cite{zhang2011coupled, wei2020multi, liu2020semantic, peng2021sparse,shi2020optimal,shrivastava2025self} have been conducted on cross-domain matching, but very few have been done in the skull-to-face matching problem. With the advancement of deep learning, we can represent 2D and 3D images in the form of feature vectors (i.e., embeddings). Following this, a study conducted by Prasad \emph{et al.}~\cite{prasad2025cross} on cross-domain skull-to-face matching with 2D skull and face images shows how deep models can be used to learn cross-domain identity representation, but this work heavily focuses on models for aligning the two modalities. A similar study was conducted on cross-domain image matching~\cite{kong2019cross}, which uses mid-level deep feature maps to match the two modalities; however, this work uses a dataset that contains impressions of shoes.

Hence, our work mainly focuses on cross-domain matching for craniofacial identification using landmarks on muilti-view 2D X-ray images of face similary for sketch images of face. Our proposed pipeline establishes a structured workflow for automatic landmark localization and skull-to-face as well as sketch-to-face matching. Our proposed framework, also perform better in sketch-to-face matching and retrieval tasks.

% Beginning with the acquisition of 204 raw skull and facial images, an elimination step is applied to remove soft tissue. The preprocessed images are then manually annotated through Roboflow, generating high-quality labeled data. These annotations are subsequently used to train YOLOv8 models for landmark localization, where YOLOv8n served as a baseline before advancing to YOLOv11n, which achieved improved accuracy through optimized training with keypoint-weighted loss. Then, 
% The enhanced YOLOv8m model demonstrated superior localization of 18 facial and 13 cranial keypoints across different orientations. Finally, the localized landmarks enable skull-to-face matching through graph convolutional networks (GCNs), effectively leveraging the structural dependencies between skull and facial attributes for craniofacial identification.

%  \begin{figure*}
%     \centering
%     \includegraphics[width =\textwidth, keepaspectratio,,trim={0.5cm 0cm 0cm 0.1cm},clip]{Images/keypoint_localization.png}
%     \caption{ Proposed pipeline for automatic landmark localization framework from raw X-ray and face images. First, boundaries of soft tissue present in raw X-ray image are manually masked and then we cropped out the soft tissue part. Then, to train the YOLOv8 model, we manually annotate the landmarks on the skull and its corresponding face image. After this YOLO model is trained over these annotated images in order to learn the landmark positions and detecting landmarks automatically.  }
%     \label{fig:framework}
% \end{figure*}

\section{Aditional Curation of S2F Dataset}\label{s2f_dataset}
\begin{figure}
    \centering
    \includegraphics[width =1\linewidth,keepaspectratio,trim={8.2cm 4.2cm 6.3cm 5.0cm},clip]{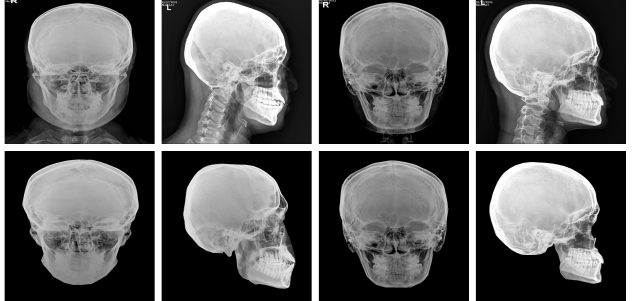}
    \caption{ Sample image of X-ray dataset used, where soft tissue eliminated images are shown down to their respective raw images. }
    \label{fig:dataset}
\end{figure}

 Our study was conducted on the S2F~\cite{prasad2025fcrinvestigatinggenerativeai} dataset which contains 51 X-ray scans with their respective face pair image from voluntary persons, aged between 21-30 years as shown in  Figure~\ref{fig:dataset}. Specifically, 22 females and 29 male volunteers are there in this dataset. More precisely, for training YOLO pose models, this dataset contains 102 skull–face pairs, which consist of 51 lateral views and 51 frontal views. We divided the dataset into training, validation, and testing sets using a 70:20:10 ratio. Specifically, 10 pairs were subject wise randomly selected for testing, 21 for validation, and 71 for training.
 
While images of the same size are usually required for batch training a neural network, the skull and face images vary in size, ranging from 153 × 258 to 819 × 1500. We employ bilinear interpolation to resize every image to a consistent 640 x 640 resolution. Then, each image is normalized by dividing its pixel values by the standard deviation of the pixel values of the data set. 

An X-ray image of the face also contains information about soft tissues, and to make the X-ray image resemble the skull image precisely, we eliminate the soft tissue part. Soft tissue elimination (STE) enables the training of a YOLO-pose~\cite{yolo_pose_v8, yolo_v11} model more precisely for skull images. This operation was applied to all the images in S2F dataset to isolate skeletal structures and facial contours. We manually outline soft tissue areas on each skull image using Roboflow's tools, specifically designed for segmentation to distinguish between soft tissue and bone. Figure~\ref{fig:dataset} shows some samples of images before and after elimination of soft tissues. [For more details on soft tissue elimination, please refer to supplementary.]
\subsection{Annotation}\label{annotation}
Annotation is the most important part of our keypoint localization system, training the YOLOv8~\cite{yolo_pose_v8} and YOLOv11~\cite{yolo_v11} pose models in detecting anatomical landmarks necessary for skull-to-face matching problem. Annotations were done by  Roboflow~\cite{roboflow2025}. For the front-facing images, 18 keypoints were manually located and annotated with labels representing significant anatomical landmarks. For side-view images, 13 keypoints were annotated. We want to outline the anatomy of the skull and face image across its length and width on the basis of these landmarks. Hence, we have chosen these landmarks to describe the morphology of skull and its corresponding face image. For example, landmark points 12, 20 and 39, 58 measure skull and face eye's anatomy across its length and width, respectively ( please refer to Figure~\ref{fig:dataset_1}).

\begin{figure*}[!ht]
    \centering
    \includegraphics[width =1.0\textwidth, keepaspectratio,trim={0.4cm 0.2cm 0.1cm 0cm},clip]{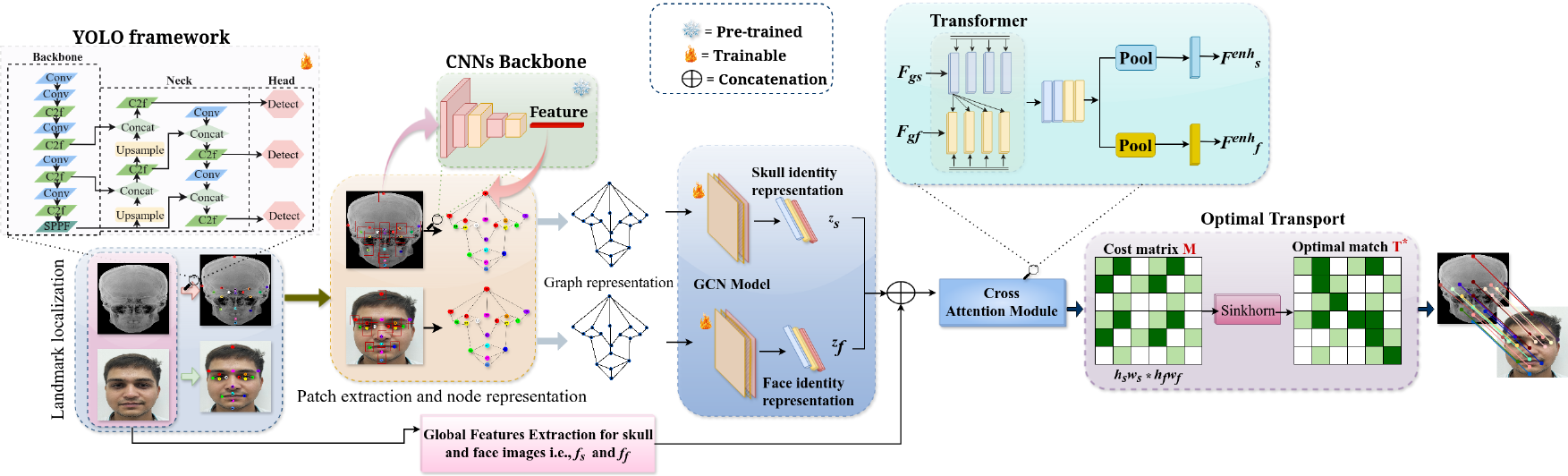}
    \caption{The proposed framework consists of five stages. In the first stage, the face and skull regions are detected, and keypoints are localized using a YOLO pose-based model. In the second stage, patches are extracted around the corresponding keypoints, and feature representations are obtained from these patches. In the third stage, each patch is treated as a node, and the connections between the nearest nodes are defined as edges, forming a graph representation. In the fourth stage, a GCNs is employed to extract high-level features from the respective graphs and this graph skull features ($z_s$) and face features ($z_f$) are then concatenated with the global features of skull ($f_{s}$) and face ($f_{f}$). Then, in the fifth stage, concatenated features of skull ($F_{gs}$) and face ($F_{gf}$) are refined through a cross-attention and optimal transport modules to establish semantic correspondence between two modalities to obtain the optimal mapping $T^{*}$.
(Best viewed in colors)}
    \label{fig:GCN_framework}
\end{figure*}

\section{Methodology}

% \subsection{\textbf{Graph Convolution Netwo-ks for feature extraction and graph matching.} }\label{GCN}

We propose a Cranio-ID framework for matching human skull images with their corresponding facial images. The framework consists of: (1) region-of-interest detection, (2) landmark localization and patch extraction, (3) local patch feature extraction and graph construction, and (4) graph embedding and global embedding extraction. (5) Feature matching using cross-attention and optimal-transport module. The overall pipeline is illustrated in Figure~\ref{fig:GCN_framework}. 

Graph structures show excellent performance at capturing structural information within the image. They can represent both nearby pixels and distant regions, which helps to show how different parts of an image are connected and related to each other. Taking advantage of these properties, we extracted features from patches around landmarks on the face and skull. In this context, each patch is represented as a node, while the connections between these patches are defined as edges. We utilize deep visual models to extract the features of these patches, which serve as the node features in the graph~\cite{sharma2025exp}. Finally, semantic correspondence between two different modalities are learned, as discussed in sections~\ref {cross_attention} to ~\ref{final_loss}. 

\subsection{Graph Representation}
Let an image $\boldsymbol{I} \in \boldsymbol{R}^{H \times W \times 3}$ contain $N$ landmarks at coordinates $\{(x_i, y_i)\}_{i=1}^{N}$ with visibility $v_i \in \{0,1\}$.
For each landmark, a square patch of size $P \times P$ is extracted. Then $I_i^\text{patch}$ patch can be represented as:
\begin{equation}
    I_i^\text{patch} = I[ x_i-P:x_i+P \; y_i-P:y_i+P]
\end{equation}
If a landmark is missing ($v_i = 0$), a zero vector is used instead. 
After this, each patch is passed through a pretrained model $\phi(\cdot)$, 
such as ResNet-18~\cite{he2016deep}, MobileNet~\cite{howard2017mobilenets}, 
EfficientNet~\cite{tan2019efficientnet}, and ViT~\cite{dosovitskiy2021an}, 
to obtain a feature vector $\mathbf{f}_i = \phi(I_i^{\text{patch}}) \in \mathbb{R}^{d_\text{feat}}$. Node features combine the landmark coordinates and the pretrained model embedding can be represented as:
\begin{equation}
    \mathbf{x}_i = [x_i, y_i, f_i] \in \mathbb{R}^{2+d_\text{feat}}
\end{equation}
Now, graph edges are formed using $k$-nearest neighbors in the 2D landmark space:
\begin{equation}
    \mathcal{N}_i = \text{argsort}_j \|\mathbf{p}_i - \mathbf{p}_j\|_2, \quad j \neq i
\end{equation}
\begin{equation}
    \mathcal{E} = \{(i,j) \mid j \in \mathcal{N}_i[1:k]\}, \quad \mathbf{p}_i=(x_i,y_i)
\end{equation}
% \subsection{Graph Generation Algorithm}

\begin{algorithm}[H]
\caption{Graph Generation from Image Landmarks}
\begin{algorithmic}[1]
\Require Image $I$, Landmark coordinates $\{(x_i, y_i, v_i)\}_{i=1}^N$, Patch size $P$, $\phi(\cdot)$
\Ensure Graph $G = (\mathbf{X}, \mathcal{E})$
\For{$i = 1$ to $N$}
    \If{$v_i == 1$}
        \State $I_i^\text{patch} \gets I[y_i-P:y_i+P, x_i-P:x_i+P]$
        \State $f_i \gets \text{$\phi(\cdot)$}(I_i^\text{patch})$
    \Else
        \State $f_i \gets \mathbf{0}$
    \EndIf
    \State $\mathbf{x}_i \gets [x_i, y_i, f_i]$
\EndFor
\State $\mathcal{E} \gets$ k-NN edges from $\mathbf{p}_i=(x_i,y_i)$
\State \Return Graph $G = (\mathbf{X}, \mathcal{E})$
\end{algorithmic}
\label{graph_generation_algo}
\end{algorithm}

\subsection{Graph Embedding via GCNs}
Given node features $\mathbf{X}$ and adjacency $\mathbf{\hat{A}}$, a GCN updates node embeddings:
\begin{equation}
    \mathbf{H}^{(0)} = \mathbf{X}, \quad
\mathbf{H}^{(l+1)} = \sigma\left( \mathbf{\hat{A}} \mathbf{H}^{(l)} \mathbf{W}^{(l)} \right)
\end{equation}

Global mean pooling produces a graph-level embedding:
\begin{equation}\label{graph_embeddings}
    \mathbf{z} = \frac{1}{N} \sum_{i=1}^{N} \mathbf{H}_i^{(L)}, \quad \mathbf{z} \in \mathbb{R}^{d_\text{embed}}
\end{equation}
% -------------------------------
% 1. Feature Extraction
% -------------------------------

\paragraph{ Global features extraction for skull and images.}
We use vision transformer (ViT) to extract the global visual skull ($f_{s} \in \mathbb{R}^{d_g}$) and face ($f_{f}\in \mathbb{R}^{ d_g}$) features from skull and face images, respectively, and $d_g$ is the ViT global feature dimension.

\paragraph{ Graph-based features.}
From Equation~\ref{graph_embeddings}, we get skull ${z}_{s}\in\mathbb{R}^{d_{embed}}$ and face graph features ${z}_{f} \in \mathbb{R}^{ d_{embed}}$,
\label{eq:graph_features} where $d_{embed}$ is the graph feature dimension.

To enrich structural information ( i.e., graph based features) with global context, the ViT global embedding is concatenated with graph embeddings.  
Thus, the fused skull (${F}_{gs}$) and face feature (${F}_{gf}$) sequences are given by
\begin{equation}
{\boldsymbol{F}}_{gs} 
= \left[\, \boldsymbol{z}_{s} \,\Vert\, \boldsymbol{f}_{s} \,\right]
\in \mathbb{R}^{d},
\label{eq:fused_skull}
\end{equation}
\begin{equation}
{\boldsymbol{F}}_{gf} 
= \left[\, \boldsymbol{z}_{f} \,\Vert\, \boldsymbol{f}_{f} \,\right]
\in \mathbb{R}^{ d}.
\label{eq:fused_face}
\end{equation}
where, $d$ = $d_{embed}$ + $d_g$ and $\vert\vert$ represents concatenation.

\subsection{Cross-Attention (CA) Based Feature Fusion}\label{cross_attention}

From Equation~\ref{eq:fused_skull} and~\ref{eq:fused_face}, $F_{gs} \in \mathbb{R}^{d}$ and $F_{gf} \in \mathbb{R}^{d}$ denote the final concatenated feature representations extracted from skull and face images, respectively. Then, fused skull and face features are projected into query, key, and value embeddings using learnable matrices $\boldsymbol{W_Q, W_K, W_V, W_Q', W_K', W_V'} \in \mathbb{R}^{d \times d}$. 
Bidirectional attention maps are then computed as:
\begin{equation}
A_{sf} = \mathrm{softmax}\left(\frac{Q_s K_f^\top}{\sqrt{d_k}}\right),
\end{equation}
\begin{equation}
A_{fs} = \mathrm{softmax}\left(\frac{Q_f K_s^\top}{\sqrt{d_k}}\right),
\end{equation}

where $Q_s = F_{gs}\boldsymbol{W_Q}$, $Q_f = F_{gf}\boldsymbol{W_Q'}$, $K_f = F_{gf}\boldsymbol{W_K}$, $K_s= F_{gs} \boldsymbol{W_K'}$, and $d_k = d / h$ for $h$-head attention. Then, the attended representations $\tilde{F}_s = A_{sf} V_f$, and $\tilde{F}_f= A_{fs} V_s$ for skull and face are refined as residual connection, layer normalization, and feed-forward networks to obtain enhanced representations $F_s^{\text{enh}}$ and $F_f^{\text{enh}}$.
% \begin{align}
% \tilde{F}_s &= A_{sf} V_f, & \tilde{F}_f &= A_{fs} V_s.
% \end{align}
where, $V_f,V_s$ are learned projections of the raw face/skull features used to transfer semantic information during cross-attention.
% \paragraph{Feature Enhancement.} Enhanced features use residual connections and layer normalization:
% \begin{align}
% F_s^{\text{enh}} &= \mathrm{LayerNorm}\!\big(F_{gs} + \tilde{F}_s\big), \\
% F_f^{\text{enh}} &= \mathrm{LayerNorm}\!\big(F_{gf} + \tilde{F}_f\big),
% \end{align}
% followed by a position-wise feed-forward network (FFN):
% \begin{align}
% F_s^{\text{enh}} &\gets \mathrm{LayerNorm}\!\big(F_s^{\text{enh}} + \mathrm{FFN}(F_s^{\text{enh}})\big), \\
% F_f^{\text{enh}} &\gets \mathrm{LayerNorm}\!\big(F_f^{\text{enh}} + \mathrm{FFN}(F_f^{\text{enh}})\big).
% \end{align}

Global embeddings for skull and face are computed  by mean pooling followed by normalization:
\begin{align}
g_s &= \mathrm{Norm}\!\left(\frac{1}{T_s} \sum_{i=1}^{T_s} F_{s,i}^{\text{enh}}\right), \\
g_f &= \mathrm{Norm}\!\left(\frac{1}{T_f} \sum_{j=1}^{T_f} F_{f,j}^{\text{enh}}\right).
\end{align}
The overall cross-attention process can be represented as:
\begin{align}
F_s^{\text{enh}} &= \mathcal{A}(F_{gs}, F_{gf}), \\
F_f^{\text{enh}} &= \mathcal{A}(F_{gf}, F_{gs}),
\end{align}
where $\mathcal{A}(\cdot)$ denotes the multi-head cross-attention operator.
This operation enables bidirectional information exchange between skull and face modalities, 
allowing each to attend to semantically corresponding regions in the other.
\subsection{Local Optimal Transport (OT)-based Alignment}

To enforce fine-grained correspondence between local skull and face regions, 
we employ an OT formulation. Given enhanced skull and face embeddings 
$F_s^{\text{enh}} \in \mathbb{R}^{d}$ and 
$F_f^{\text{enh}} \in \mathbb{R}^{d}$, 
we first compute the pairwise cost matrix $\boldsymbol{C_{sf}} \in \mathbb{R}^{d \times d}$ using cosine similarity:
\begin{equation}
C_{sf}(i,j) = 1 - \frac{(F_s^{\text{enh}})_i \cdot (F_f^{\text{enh}})_j}{\|(F_s^{\text{enh}})_i\|_2 \, \|(F_f^{\text{enh}})_j\|_2}.
\end{equation}

The entropic-regularized optimal transport problem is then formulated as:
\begin{align}
T^* &= \arg \min_{T \in \Pi(\mu,\nu)} \langle T, C_{sf} \rangle - \varepsilon H(T), \\
\Pi(\mu, \nu) &= \{ T \ge 0 \mid T \mathbf{1}_{N_f} = \mu, \, T^\top \mathbf{1}_{N_s} = \nu \}, \\
H(T) &= - \sum_{i,j} T_{ij} \log T_{ij}.
\end{align}

OT-based similarity between skull and face embeddings is then defined as:
\begin{equation}
\mathcal{S}_{\text{OT}}(F_s^{\text{enh}}, F_f^{\text{enh}}) = - \langle T^*, C_{sf} \rangle = - \sum_{i,j} T_{ij}^* C_{sf}(i,j),
\end{equation}
and accordingly, the OT loss is defined as:
\begin{equation}
\mathcal{L}_{\text{OT}} = \langle T^*, C_{sf} \rangle.
\end{equation}
which penalizes the transport cost between corresponding local features.

\subsection{Combined Global and OT-based Triplet Loss}\label{final_loss}

\paragraph{Global Similarity.} For skull-face pairs:
\begin{equation}
S_{ij} = \cos(g_{s,i}, g_{f,j}) = \frac{g_{s,i}^\top g_{f,j}}{\|g_{s,i}\|_2 \, \|g_{f,j}\|_2}.
\end{equation}

\paragraph{Combined Similarity for Triplet.} Using a weighting factor $\beta \in [0,1]$:
\begin{align}
\text{sim}^{\text{pos}}_i &= \beta \, S_{ii} + (1-\beta) \, \tanh(\mathcal{S}_{\text{OT}}^{\text{pos}})_i, \\
\text{sim}^{\text{neg}}_i &= \beta \, S_{i j_i^-} + (1-\beta) \, \tanh(\mathcal{S}_{\text{OT}}^{\text{neg}})_i,
\end{align}
 where $S_{ii}$ is the positive global similarity, $S_{i j_i^-}$ is the hardest negative, and $\mathcal{S}_{\text{OT}}^{\text{pos}}$, $\mathcal{S}_{\text{OT}}^{\text{neg}}$ are OT similarities.
\paragraph{Triplet Loss.} The final triplet loss combining global and OT-based similarities:

\begin{equation}
\label{triplet_loss}
\mathcal{L}_{\text{triplet}} = \frac{1}{B} \sum_{i=1}^{B} \max \Big( 0, \, m - \text{sim}^{\text{pos}}_i + \text{sim}^{\text{neg}}_i \Big),
\end{equation}
with margin $m$.

\paragraph{Total Training Objective.} Including the OT cost as auxiliary:
\begin{equation}
\mathcal{L}_{\text{total}} = \mathcal{L}_{\text{triplet}} + \lambda_{\text{OT}} \, \mathcal{L}_{\text{OT}},
\end{equation}
where $\lambda_{\text{OT}}$ balances local OT alignment. This joint formulation ensures that the model learns globally discriminative and locally aligned cross-domain representations.

% \begin{table}[!t]
% \caption{Quantitative performance comparison of YOLOv8n and YOLOv11n across \textbf{CUHK Face-Sketch}~\cite{zhang2011coupled} dataset types. 
% B = bounding box metrics, P = pose estimation metrics.}
%\label{tab:yolo_style_matched_CUHK}
%\centering
%{\large
%\resizebox{\linewidth}{!}{ % or \linewidth
%\begin{tabular}{l|cc|cc}
% \hline
% \textbf{Model (Dataset)} 
% \multicolumn{2}{c}{\textit{CUHK Face-Sketch}} \\
% \midrule
% & \multicolumn{2}{c|}{\textbf{Bounding Box (B)}} 
% & \multicolumn{2}{c}{\textbf{Pose (P)}} \\ 
% \cline{2-5}
%  & \textbf{mAP50 ↑} & \textbf{mAP50-95 ↑}
%   & \textbf{mAP50 ↑} & \textbf{mAP50-95 ↑} \\ 
% \hline
% YOLOv8n (Face)        & 99.5 & 93.4  & 99.5 & 99.4 \\
% YOLOv11n (Face)       & 99.5 & 97.6  & 99.5& 99.4 \\
% YOLOv8n (Sketch)       & 99.5& 82.4 & 99.5& 86.7 \\
% YOLOv11n (Sketch)      & 99.5& 89.5 & 99.5& 75.2 \\
% YOLOv8n (Face+Sketch)  & 99.5 & 88.6  & 99.5& 94.2 \\
% YOLOv11n (Face+Sketch) & 99.5& 92.2  & 99.5 & 85.1 \\ 
% \hline
% \end{tabular}
% }}
% \end{table}

% \begin{figure}
%     \centering
%     \includegraphics[width=9cm,keepaspectratio,trim={3.2cm 2cm 1cm 1.1cm},clip]{Images/mAP.pdf}
%     \caption{ (a) shows mAP50 and (b) shows mAP50-95 for combined skull and face dataset on YOLOv11n pose model over 400 epochs. }
%     \label{fig:map}
% \end{figure}

\section{Experiments}
To demonstrate the effectiveness of our method, we conducted tests using two publicly available datasets: S2F~\cite{prasad2025fcrinvestigatinggenerativeai} and CUHK~\cite{zhang2011coupled}. The reason behind choosing CUHK datset is that it is more domain wise relevant with S2F dataset. In simple words, skull image in S2F is more correlated with sketch image in CUHK and face image in S2F is highly correlated with face images in CUHK.  We evaluate our results using two key metrics: Recall at K (R@K), mean Average Precision at K (mAP@K). We also conduct ablation studies to take a closer look at how our method works on different values of hyperparameters (i.e, patch dimension $d$ and margin $m$).
% In this study, three YOLO-based pose estimation models~\cite{ultralytics2023yolov8pose}, \cite{ultralytics2025yolov11} were trained using the Ultralytics framework~\cite{ultralyticsgithub2025} with pretrained YOLOPose weights. The first experiment focused on facial detection and landmark localization. The dataset included both front-view and side-view facial images, annotated with 18 landmarks for the front face and 13 landmarks for the side face. 

\subsection{Implementation Details}
The detection and landmark localization models were trained with the same hyperparameter configuration. Training images were resized to 640 × 640 pixels, and a batch size of 8 was used. Each model was trained for up to 500 epochs, with early stopping applied after 50 stagnant epochs to prevent overfitting. Optimization was handled automatically by the Ultralytics training pipeline, using an initial learning rate of 0.01, momentum of 0.937, and weight decay of 0.0005, with a three-epoch warm-up schedule. To improve generalization, multiple augmentation strategies were applied, including random translation (10\%), scaling (up to 50\%), horizontal flipping (50\%).

For each landmark, three different patch sizes i.e., $32 \times 32$,  $64 \times 64$ and  $128 \times 128$ was extracted and passed through pretrained CNNs (ResNet18, MobileNetV2, EfficientNet-B0) and ViT-Base patch16 to obtain feature embeddings, which were concatenated with the landmark coordinates to form node features. Graphs were constructed using algorithm~\ref{graph_generation_algo}. A two-layer GCNs with hidden and embedding dimensions of 128 is then used to extract the embeddings of these input graphs. 
% When we get embeddings of graphs from GCN, then we concatenate the global skull and face features with these graph features. These concatenated features are then passed to two different modules: one is cross-attention, and the other is optimal transport to find the semantic correspondence between the embeddings of two different modalities.

Our proposed framework with CA-OT model is trained for 50 epochs with the best hyperparameter configurations: learning rate 0.0001, batch size 16, margin (i.e., \emph{m} = 0.3) for triplet loss, weight decay 0.00001, sinkhorn iterations 80, and number of heads in attention 4.
We have maintained the same number of iterations for training the S2F and CUHK datasets. To tackle the overfitting, we choose the snapshot of the model based on the validation set.

\begin{table}[!t]
\caption{Quantitative comparison of YOLOv8n and YOLOv11n on landmark localization across \textbf{S2F}~\cite{prasad2025fcrinvestigatinggenerativeai} and \textbf{CUHK Face-Sketch}~\cite{zhang2011coupled}. 
B = bounding box metrics, P = pose estimation metrics.}
\label{tab:yolo_style_combined}
\centering
\large
\resizebox{\linewidth}{!}{
\begin{tabular}{l|cc|cc}
\hline
\textbf{Model} 
& \multicolumn{2}{c|}{\textbf{Bounding Box (B)}} 
& \multicolumn{2}{c}{\textbf{Pose (P)}} 
\\
\cline{2-5}
 & \textbf{mAP50 ↑} & \textbf{mAP50–95 ↑}
 & \textbf{mAP50 ↑} & \textbf{mAP50–95 ↑}
\\
\hline

\multicolumn{5}{c}{\textit{\textbf{S2F Dataset}}} \\
\midrule
YOLOv8n (Face)        & 99.5 & 88.1 & 99.5 & 94.0 \\
YOLOv11n (Face)       & 99.5 & 87.3 & 99.5 & 80.9 \\
YOLOv8n (Skull)       & 99.5 & 98.0 & 99.5 & 94.5 \\
YOLOv11n (Skull)      & 99.5 & 58.9 & 99.5 & 85.1 \\
YOLOv8n (Face+Skull)  & 99.5 & 92.8 & 99.5 & 88.6 \\
YOLOv11n (Face+Skull) & 93.7 & 95.8 & 92.3 & 92.3 \\
\hline

\multicolumn{5}{c}{\textit{\textbf{CUHK Face-Sketch Dataset}}} \\
\midrule
YOLOv8n (Face)         & 99.5 & 93.4 & 99.5 & 99.4 \\
YOLOv11n (Face)        & 99.5 & 97.6 & 99.5 & 99.4 \\
YOLOv8n (Sketch)       & 99.5 & 82.4 & 99.5 & 86.7 \\
YOLOv11n (Sketch)      & 99.5 & 89.5 & 99.5 & 75.2 \\
YOLOv8n (Face+Sketch)  & 99.5 & 88.6 & 99.5 & 94.2 \\
YOLOv11n (Face+Sketch) & 99.5 & 92.2 & 99.5 & 85.1 \\
\hline
\end{tabular}
}
\end{table}

\subsection{Performance Comparison}
In our study, we conducted a comparison of various deep learning models, including transformer-based architectures, utilizing the S2F and CUHK datasets. Our main objective is to map skull-to-face, but to show the applicability and generalization of our method, we have also conducted experiments on CUHK dataset. The results of this comparison are presented in Table~\ref{tab:yolo_style_combined}, \ref{tab:shared_backbone_s2f_cuhk}, \ref{tab:ablation_studies_combined}, \ref{tab:ablation_margin_combined}.

\begin{table*}[t]
\centering
\caption{
\textbf{Cross-domain retrieval results on the S2F~\cite{prasad2025cross} and CUHK~\cite{zhang2011coupled} dataset.}  
Comparative quantitative result with metrics Recall@K, mAP@K (\%). Models combining both Optimal Transport (OT) and Cross-Attention (CA) yield consistent gains across backbones ( d = 128 and m = 0.3). Best values are marked in Bold.}
\label{tab:shared_backbone_s2f_cuhk}
\setlength{\tabcolsep}{4pt}
\renewcommand{\arraystretch}{1.05}
\small

\resizebox{\textwidth}{!}{
\begin{tabular}{lcc|cccc|cccc|cccc|cccc}
\toprule
\multirow{3}{*}{\textbf{Backbone}} &
\multirow{3}{*}{\textbf{OT}} &
\multirow{3}{*}{\textbf{CA}} &

\multicolumn{8}{c|}{\textbf{S2F Dataset}} &
\multicolumn{8}{c}{\textbf{CUHK Dataset}} \\

\cmidrule(lr){4-11} \cmidrule(lr){12-19}
 & & &
\multicolumn{4}{c|}{\textbf{Recall@K (\%)}} &
\multicolumn{4}{c|}{\textbf{mAP@K (\%)}} &
\multicolumn{4}{c|}{\textbf{Recall@K (\%)}} &
\multicolumn{4}{c}{\textbf{mAP@K (\%)}} \\

\cmidrule(lr){4-7} \cmidrule(lr){8-11}
\cmidrule(lr){12-15} \cmidrule(lr){16-19}

 & & &
R@1 & R@5 & R@10 & R@20 &
mAP@1 & mAP@5 & mAP@10 & mAP@20 &
R@1 & R@5 & R@10 & R@20 &
mAP@1 & mAP@5 & mAP@10 & mAP@20 \\
\midrule

% ========= CNN MODELS =========
\multicolumn{19}{c}{\textit{CNN-based Models}} \\
\midrule

% ---------------- RESNET-18 ------------------
\multirow{4}{*}{ResNet-18}
 & \ding{55} & \ding{55} & 36.9 & 49.4 &52.2 & 60.2 & 36.9 & 42.8 & 43.3& 44.3 & 75.9 & 86.3 & 89.5 & 90.7 & 75.9 & 80.2 & 80.8 & 81.1 \\
 & \ding{55} & \ding{51} & 50.0 & 61.9 & 69.8 & 82.3 & 50.0 & 57.8 & 59.9 & 61.6 & 88.4 & 89.0 & 89.3 & 90.8 & 88.4 & 88.8 & 88.9 &89.3\\
 & \ding{51} & \ding{55} & 36.9 & 49.4 &52.2 & 60.2 & 36.9 & 42.8 & 43.3& 44.3 & 75.9 & 87.6 & 89.6 & 90.7 & 75.9 & 80.5 & 81.0 & 81.2 \\
 & \ding{51} & \ding{51} &
 \textbf{50.0} & \textbf{71.5} & \textbf{78.9} & \textbf{86.9} &
 50.0 & 65.2& 67.3 & 68.3 &
 \textbf{88.4} & \textbf{89.0} & \textbf{89.3} & 90.3&
 88.5& 88.9 & 89.0 & 89.2 \\
\midrule

% ---------------- MOBILENET ------------------
\multirow{4}{*}{MobileNet}
 & \ding{55} & \ding{55} & 36.3 & 48.3 & 52.8 & 59.6 & 36.3 & 42.0 & 42.9& 43.8 & 75.9 & 86.3 & 89.6 & 90.7 & 75.9 & 80.4 & 81.0 & 81.3 \\
 & \ding{55} & \ding{51} & 50.0 & 60.8 & 69.3 & 82.9 & 50.0 & 57.5 & 59.8 & 61.7 & 88.4 & 89.0 & 89.5 & 90.5 & 88.4 & 88.7 & 88.9 & 89.2 \\
 & \ding{51} & \ding{55} & 36.9 & 48.8 & 53.4 & 59.0 & 36.9 & 42.7 & 43.5 & 44.3 & 75.9& 87.5 & 89.6 & 90.7 & 75.9 & 80.4 & 80.9 & 81.2 \\
 & \ding{51} & \ding{51} &
 \textbf{50.0} & \textbf{73.3} & \textbf{80.1} & \textbf{85.8} &
 50.0& 65.8 & 67.7 & 68.5 &
 \textbf{88.4} & \textbf{89.0} & 89.3 & 90.3 &
 88.5 & 88.8 & 89.0 & 89.2 \\
\midrule

% ---------------- EFFICIENTNET ------------------
\multirow{4}{*}{EfficientNet}
 & \ding{55} & \ding{55} & 35.8 & 48.8 & 52.8 &59.0 & 35.8 & 42.0 & 42.7 & 43.6 & 75.9 & 86.4 & 89.6 & 90.7 & 75.9 & 80.4 & 81.0 & 81.3 \\
 & \ding{55} & \ding{51} & 50.0 & 63.0 & 71.5 & 83.5 & 50.0 & 58.6 & 60.7 & 62.4 & 88.4 & 89.0 & 89.5 & 90.8 &88.4 & 88.8 & 89.0 & 89.3 \\
 & \ding{51} & \ding{55} & 36.3 & 48.8 & 53.4 &60.8 & 36.3 & 42.2 & 43.1 & 44.0 &  75.9 & 87.5 & 89.5 & 90.7 & 75.9 & 80.4 & 80.8 & 81.2 \\
 & \ding{51} & \ding{51} &
 \textbf{50.0} & \textbf{73.3} & \textbf{80.6} & \textbf{86.3} &
 50.0 & 66.1 & 68.0 &68.8&
 \textbf{88.4} & \textbf{89.0} & 89.3 & 90.4 &
88.4 & 88.3 & 88.9 & 89.1 \\
\midrule

% ========= TRANSFORMER MODELS =========
\multicolumn{19}{c}{\textit{Transformer-based Model}} \\
\midrule

\multirow{4}{*}{ViT}
 & \ding{55} & \ding{55} &
 37.5 & 48.3 &51.7 & 60.2 &
 37.5 & 42.6 & 43.3 & 44.4 &
75.9 & 87.6 & 89.6 & 90.7 &
 75.9 & 80.5 & 80.9 & 81.2 \\
  
 & \ding{55} & \ding{51} &
 50.0 & 63.0 & 72.7 & 83.5 &
 50.0 & 58.7 & 61.2 & 62.8 &
 88.4 &89.0  & 89.3 &90.5&
 88.4 & 88.7 & 88.9 & 89.2 \\

 & \ding{51} & \ding{55} &
 35.8 & 47.7 & 52.2 & 59.0 &
 35.8 & 41.5 & 42.4 & 43.3 &
 75.9 &87.6  &89.6 & 90.7 &
75.9 &80.5  &80.9 &81.2  \\

 & \ding{51} & \ding{51} &
 \textbf{50.0} & \textbf{73.3} & \textbf{78.4} & \textbf{85.8} &
 \textbf{50.0}&\textbf{66.6}& \textbf{68.1} & \textbf{69.1}&
 \textbf{88.4} & \textbf{89.0} & \textbf{89.3} & \textbf{90.6} &
 88.4 & 88.8& 88.9 & 89.2\\
\bottomrule
\end{tabular}
}
\end{table*}
\begin{figure}[!t]
    \centering
    \includegraphics[width =1\linewidth,keepaspectratio,trim={7.5cm 0.9cm 6.7cm 1cm},clip]{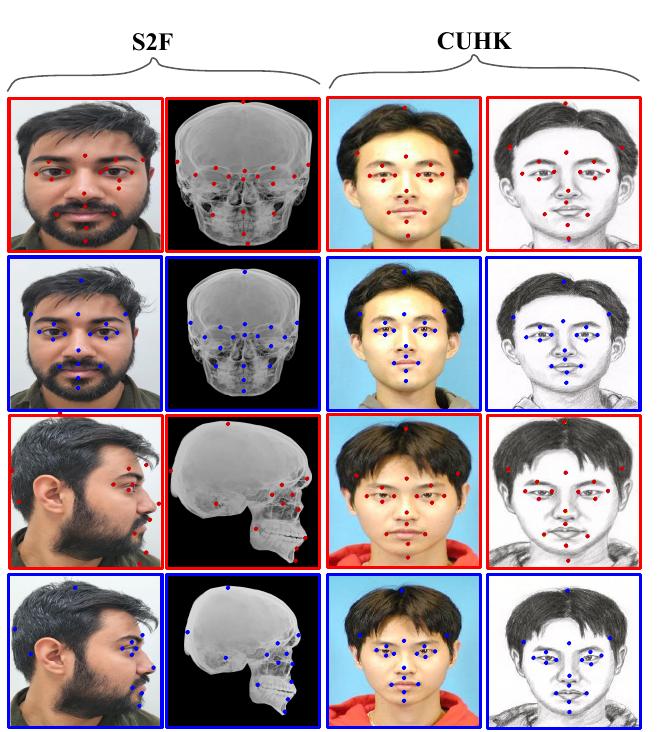}
    \caption{ Predictions on test dataset samples for landmark localization on S2F dataset and CUHK dataset. Where red one represents the prediction results while blue represents the ground truth. }
    \label{fig:landmark_prediction}
\end{figure}
\begin{figure}[!t]
    \centering
    \includegraphics[width =1\linewidth,keepaspectratio,trim={3.5cm 3cm 2.5cm 2.0cm},clip]{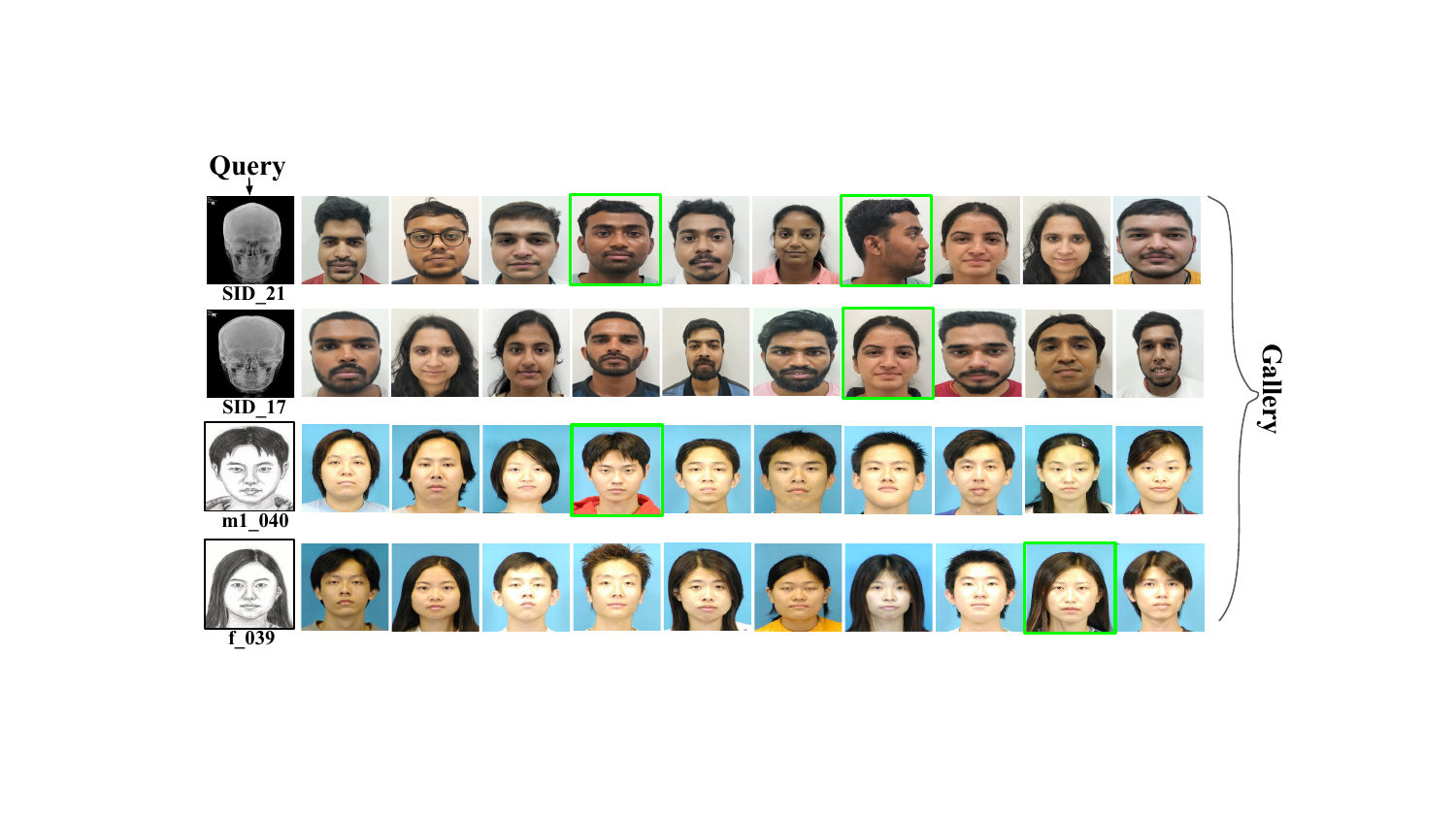}
    \caption{Top-10 retrieval results for given query in S2F dataset and CUHK dataset. The green box represents a correct match for the given query image (i.e, skull or sketch). }
    \label{fig:retrieval}
\end{figure}
\textbf{S2F dataset:} Table~\ref{tab:yolo_style_combined} presents quantitative results on landmark detection and location on S2F dataset. The performance of YOLO8n and YOLO11n models was evaluated across three datasets: face, skull, and a combined face+skull dataset, using standard object detection metrics including mean Average Precision (mAP) at IoU thresholds of 0.5 and 0.5:0.95. Overall, YOLO8n is more consistent performance across datasets, whereas YOLO11n achieved competitive accuracy with proposed dataset. Table ~\ref{tab:shared_backbone_s2f_cuhk} presents the quantitative results on S2F dataset, showing the effectiveness of each module, where our proposed method outperforms all other methods for skull-to-face image retrieval, achieving 50.0(\%), 73.3(\%), 78.4(\%) and 85.8(\%) for R@1, R@5, R@10 and R@20 with our best model configuration having backbone ViT base 16. This table also presents results on mAP@k.
% YOLO8n demonstrated consistently high performance, achieving near-perfect mAP@0.5 values of 0.995 for each dataset. While the mAP@0.5:0.95 metric was slightly lower for the face+skull dataset (0.9279 for bounding box evaluation), overall performance remained robust. YOLO11n, a more recent architecture, similarly achieved perfect recall across all datasets. While mAP@0.5 remained high for individual datasets (0.995), the Face+Skull dataset showed reduced performance (0.937). The mAP@0.5:0.95 metric showed greater variability, particularly for the Skull dataset (0.5885), indicating less stability under stricter IoU thresholds. 

\textbf{CUHK dataset:} Table~\ref{tab:yolo_style_combined} shows quantitative results on landmark localization on CUHK dataset.
Table~\ref{tab:shared_backbone_s2f_cuhk} presents the quantitative results on CUHK dataset, showing the effectiveness of each module, where our proposed method outperforms all other methods sketch-to-face image retrieval, achieving 88.4(\%), 89.0(\%), 89.3(\%) and 90.6(\%) for R@1, R@5, R@10 and R@20. This table also presents results on mAP@k. 
\begin{figure*}[!t]
    \centering
    \includegraphics[width=15cm,keepaspectratio,trim={1.8cm 2.5cm 3cm 2cm},clip]{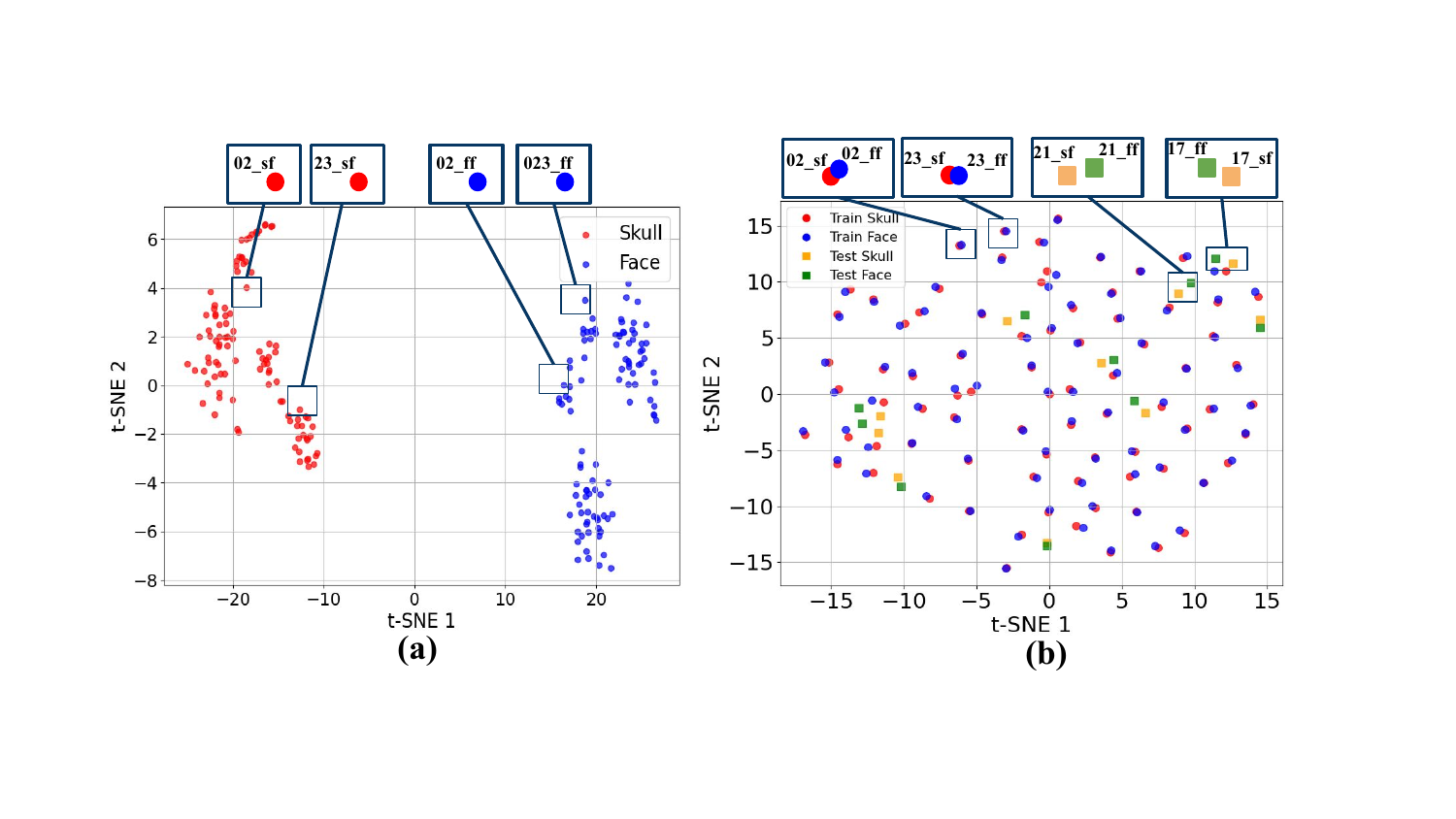}
    \caption{(a) 2D visualisation of embeddings before training shows a large domain gap between two modalities (i.e., skull and face). A few skull (red circle) and face (blue circle) embeddings are shown in boxes. While (b) shows train and test embeddings after training, indicating how two different modalities are overlapped after training of our proposed framework (sf represents front skull and ff represents front face). (Best viewed in colors)}
    \label{fig:_train_test_tsne}
\end{figure*}
Hence, from Table~\ref{tab:shared_backbone_s2f_cuhk} we can conclude that finding semantic correspondence between skull-to-face in S2F dataset is more challenging than sketch-to-face in CUHK dataset due to large domain gap between skull and face image compared to sketch and face images. That's why retrieval metrics for the CUHK dataset is better than those for the S2F dataset.
 Figure~\ref{fig:landmark_prediction} shows the prediction result on the S2F and CUHK dataset, where the frontal face, skull and sketch have better localization of the landmarks, but in the side skull and face, localization of landmarks are not very accurate. 

\begin{table}[htbp]
\centering
\caption{Impact of hidden space dimensionality ($d$), corresponding to the patch size per keypoint, for both datasets using our best performing ViT model ($m = 0.3$) with OT + Cross-Attention.}
\label{tab:ablation_studies_combined}
\footnotesize
\begin{tabular}{lccccc}
\toprule
\textbf{Dataset} & \textbf{$d$} & \textbf{R@1} & \textbf{R@5} & \textbf{R@10} & \textbf{R@20} \\
\midrule
\multirow{3}{*}{S2F} 
& 32  & 49.4 & 62.5 & 70.4 & 80.6 \\
& 64  & 48.8 & 63.0 & 71.0 & 80.6 \\
& 128 & \textbf{50.0} & \textbf{73.3} & \textbf{78.4} & \textbf{85.8} \\
\midrule
\multirow{3}{*}{CUHK} 
& 32  & 88.4 & 89.0 & 89.2 & 90.5 \\
& 64  & 88.4 & 88.6 & 89.3 & 90.4 \\
& 128 & \textbf{88.4} & \textbf{89.0} & \textbf{89.3} & \textbf{90.6} \\
\bottomrule
\end{tabular}
\end{table}

\subsection{Ablation studies}
Ablation studies are conducted on S2F and CUHK dataset testing set to evaluate the effectiveness of different dimensions of patch size around the landmarks. Table~\ref{tab:ablation_studies_combined} shows how different patch sizes ( i.e., 32, 64, 128) impact the performance of our model, and it can be seen that the best result is when the dimension of the patch size is 128. We also check the effectiveness of different values of hyperparameter \emph{m} in equation~\ref{triplet_loss}. Table~\ref{tab:ablation_margin_combined} shows how this hyperparameter can impact on the performance of our model. We obtain best performance when \emph{m} is set to 0.3.
Figure~\ref{fig:retrieval} shows some retrieval results on S2F and CUHK datasets showing the effectiveness of our proposed model in retrieving faces from the gallery for given query images of skulls and sketches. Figure~\ref{fig:_train_test_tsne} shows a large modality gap between the skull and face embeddings in the S2F dataset because skull visual attributes are very low aligned with facial visual attributes. Additionally, this figure also shows how embeddings from these two different modalities are overlapped after training of our proposed model. 
\begin{table}[htbp]
\centering
\caption{Ablation study on margin ($m$) for both S2F (Skull-to-Face) and CUHK (Sketch-to-Face) datasets using ViT ($d = 128$) with OT + Cross-Attention.}
\label{tab:ablation_margin_combined}
\footnotesize
\begin{tabular}{p{2.25cm}ccccc}
\toprule
\textbf{Dataset} & \textbf{$m$} & \textbf{R@1} & \textbf{R@5} & \textbf{R@10} & \textbf{R@20} \\
\midrule
\multirow{4}{*}{S2F} 
& 0.1 & 46.0 & 53.4 & 59.0 & 68.1 \\
& 0.2 & 46.0 & 60.8 & 66.4 & 78.4 \\
& 0.3 & \textbf{50.0} & \textbf{73.3} & \textbf{78.4} & \textbf{85.8} \\
& 0.4 & 49.4 & 61.3 & 68.7 & 81.8 \\
\midrule
\multirow{4}{*}{CUHK} 
& 0.1 & 88.4 & 88.7 & \textbf{89.6} & \textbf{90.7} \\
& 0.2 & 88.4 & 88.9 & 89.3 & 90.5 \\
& 0.3 & \textbf{88.4} & \textbf{89.0} & 89.3 & 90.6 \\
& 0.4 & 88.4 & 88.6 & 89.1 & 90.2 \\
\bottomrule
\end{tabular}
\end{table}
Hence, our proposed work, Cranio-ID, effectively aligns cross-domain modalities, enabling more accurate and reliable craniofacial identification and retrieval.
[For more details and understanding, please refer to the supplementary.]

\section{Conclusions}
In this paper, we introduced a framework Cranio-ID for automatically annotating 19 craniofacial landmarks on 2D images of skulls and faces, including both frontal and lateral views. Our proposed method accurately identifies these landmarks, providing an alternative to manual annotation. We systematically studied the impact of our approach on cross-domain image matching, specifically focusing on skull-to-face matching. Our results indicate that our proposed method performs significantly better in sketch-to-face matching, demonstrating its general applicability in various matching scenarios. Future work will explore how landmark-based methods can be utilized for craniofacial reconstruction. Overall, our results highlight the effectiveness of the proposed framework, establishing it as a valuable tool in forensic investigations.

{
    \small
    \bibliographystyle{ieeenat_fullname}
    \bibliography{main}
}

\end{document}